\documentclass[runningheads]{llncs}
\usepackage[T1]{fontenc}

\title{Sequential, parallel and consecutive hybrid evolutionary-swarm optimization metaheuristics}
\titlerunning{Sequential, parallel and consecutive hybrid evolutionary-swarm\ldots}

\author{Piotr Urbańczyk\inst{2,1}\orcidID{0000-0001-8838-2354}
\and Aleksandra Urbańczyk\inst{1}\orcidID{0000-0002-6040-554X}
\and Magdalena Król\inst{1}\orcidID{0000-0003-0392-0921}
\and Leszek Rutkowski\inst{3}\orcidID{0000-0001-6960-9525}
\and Marek Kisiel-Dorohinicki\inst{1}\orcidID{0000-0002-8459-1877}}

\authorrunning{P. Urbańczyk et al.}

\institute{AGH University of Krakow, Al. Mickiewicza 30, 30-059 Krakow, Poland
\email{\{purbanczyk,aurbanczyk,magdakrol,doroh\}@agh.edu.pl} \and
Jagiellonian University, ul. Gołębia 24, 31-007 Kraków, Poland
\email{piotr.urbanczyk@uj.edu.pl}
\and
Systems Research Institute, Polish Academy of Sciences,
6 Newelska Street, 01-447 Warsaw, Poland
\email{rutkowski@agh.edu.pl}
}

\usepackage[dvipsnames]{xcolor}

\usepackage{amsmath}
\usepackage[nopatch]{microtype}
\usepackage{booktabs}

\usepackage{float}
\usepackage[ruled,linesnumbered,lined]{algorithm2e}
\usepackage{algpseudocode}
\usepackage{subcaption}
\usepackage{makecell}
\usepackage{multirow}
\usepackage{multicol}
\usepackage{url,xcolor}

\usepackage{threeparttable}
\usepackage{tabularx}
\newcolumntype{s}{>{\hsize=.6\hsize}X}
\newcolumntype{k}{>{\hsize=.7\hsize}X}
\usepackage{rotating}
\usepackage{array}
\usepackage{makecell}
\usepackage{pbox}
\usepackage[hidelinks]{hyperref}

\begin{document}
\allowdisplaybreaks

\maketitle
\begin{abstract}
The goal of this paper is twofold. First, it explores hybrid evolutionary-swarm metaheuristics that combine the features of PSO and GA in a sequential, parallel and consecutive manner in comparison with their standard basic form: Genetic Algorithm and Particle Swarm Optimization. The algorithms were tested on a set of benchmark functions, including Ackley, Griewank, Levy, Michalewicz, Rastrigin, Schwefel, and Shifted Rotated Weierstrass, across multiple dimensions. The experimental results demonstrate that the hybrid approaches achieve superior convergence and consistency, especially in higher-dimensional search spaces.
The second goal of this paper is to introduce a novel consecutive hybrid PSO-GA evolutionary algorithm that ensures continuity between PSO and GA steps through explicit information transfer mechanisms, specifically by modifying GA's variation operators to inherit velocity and personal best information.


\end{abstract}

\section{Introduction}\label{sec:intro}
While numerous hybrid algorithms integrating Particle Swarm Optimization (PSO) and Genetic Algorithms (GA) have been proposed (as reviewed, e.g., in  \cite{thangaraj2011particle,placzkiewicz2018hybrid,sengupta2019particle,Shao2023PGA}), this work focuses on a systematic comparison of three distinct hybridization modes---sequential, parallel, and a tightly coupled consecutive approach---across a wide range of problem dimensions. To achieve this, we introduce and evaluate a specific implementation of the consecutive approach (PGCHEA) featuring a novel mechanism for preserving search momentum (velocity and personal bests) across the alternating GA steps, addressing a potential information loss in simpler sequential hybrids. We specifically investigate how these different strategies handle the challenges posed by increasing dimensionality.


\section{Hybrid models of evolutionary and swarm optimization methods}
This section briefly introduces the algorithms we have investigated in our research, along with referring to the relevant papers showing the hybrid versions of metaheuristics under consideration.
\enlargethispage{2.5\baselineskip}

\subsection{Genetic Algorithm (GA)}\label{sec:ga}
Genetic Algorithms are one of the earliest and perhaps most common computational models inspired by nature and evolution. They were originally developed by John Holland \cite{holland1975adaptation} and then later modified by Goldberg \cite{goldberg1989genetic}. These population-based algorithms encode potential solutions to specific problems as chromosome-like data structures and apply stochastic genetic search operators to these structures to preserve and mute the ``genes'' they carry. In the general form, the method can be described as:
\begin{equation}\label{eq:ga}
    P' = \rho \bigl(P \cup \mu\left(\sigma\left(P, f\right)\right)\bigr)
\end{equation}
where $P$ is a multiset of positions in the search space (solution candidates), called the population, $f$ is a fitness
function that evaluates the solution candidates and returns a vector of values that stand for the optimality of each population
member, $\mu$ is a composite operator that introduces random variations to a subset of individuals within the population, $\sigma$ is a stochastic selection operator, and \(\rho\)~denotes the replacement strategy that removes poorly performing individuals.


In our implementation, selection \(\sigma\) uses binary tournament: two individuals are randomly drawn, and the fitter one is added to the mating pool. The variation operator \(\mu\) applies simulated binary crossover (SBX) and polynomial mutation. SBX recombines two real-valued parents into offspring, while mutation adds diversity by perturbing gene values. These operators generate a new population aimed at improving solution quality. The process repeats until a termination criterion—such as a maximum number of iterations or a desired fitness level—is met. The best individual from the final population is returned as the solution. The algorithm proceeds as follows:

\begin{algorithm}[H]
\caption{GA}
\KwIn{Population size $N$, crossover rate $p_c$, mutation rate $p_m$}
\KwOut{Best solution found}
Initialize population $P$ with $N$ individuals\;
Evaluate fitness of each individual in $P$\;
\While{Termination criterion is not met}{
    Select parents from population $P$ based on their fitness\;
    Apply crossover to selected parents with probability $p_c$ to produce offspring\;
    Apply mutation to offspring with probability $p_m$\;
    Evaluate fitness of offspring\;
    Replace less fit individuals in $P$ with offspring to create new population\;
}
\Return the best individual from the final population\;
\end{algorithm}

\subsection{Particle Swarm Optimization (PSO)}\label{sec:pso}
\nocite{eberhart1995new}
Particle Swarm Optimization (PSO) is an evolutionary computational model inspired by the social behavior of swarms, such as flocks of birds or schools of fish. Introduced by Kennedy and Eberhart \cite{kennedy1995particle}, PSO is a population-based optimization method similar to Genetic Algorithms (GA). The algorithm begins with a set of randomly generated potential solutions, termed particles, which collectively form the initial population, often termed swarm. In general, the PSO method can be described as:
\begin{equation}\label{eq:pso}
P' = m(P, f)
\end{equation}
where \(P\) represents a multiset of positions, \(f\) is the fitness function that evaluates each particle, and \(m\) is a stochastic population manipulation function that produces a new population from the current one.

In a \(n\)-dimensional search space, each particle is represented by the position vector \(X_i = (x_{i1}, x_{i2}, \dots, x_{in})\). The particle also has an associated velocity vector \(V_i = (v_{i1}, v_{i2}, \dots, v_{in})\). The fitness of each particle is evaluated using an objective function, with higher fitness values indicating better solutions. Each particle retains a personal best position \(p_i\) and is influenced by the global best position \(p_g\) discovered by the entire swarm.

The movement of the particles is guided by the following equations:
\begin{equation}\label{eq:velocity_update}
v_{ij}' = w v_{ij} + c_1 r_1 (p_{ij} - x_{ij}) + c_2 r_2 (p_{gj} - x_{ij})
\end{equation}
\begin{equation}\label{eq:position_update}
x_{ij}' = x_{ij} + v_{ij}',
\end{equation}
where \(c_1\) and \(c_2\) are acceleration coefficients (sometimes called \textit{cognitive} and \textit{social} coefficients, respectively), \(r_1\) and \(r_2\) are random values uniformly distributed in the range [0, 1], and $j=1,\ldots,n$. The velocity equation directs a particle’s movement toward both its personal best and the global best positions, encouraging convergence to an optimal solution over successive iterations.

This iterative process of evaluating fitness, updating velocities and positions, and adjusting towards the best solutions continues until a termination criterion is met. The best position found by any particle at the end of the process is returned as the final solution.

\begin{algorithm}[H]
\caption{PSO}
\KwIn{Swarm size \(N\), acceleration coefficients \(c_1\) and \(c_2\), inertia weight \(w\)}
\KwOut{Best solution found}
Initialize population of \(N\) particles with random velocities\;
Evaluate the fitness of each particle\;
    \While{Termination criterion is not met}{
        Modify each particle's position and velocity by equations (\ref{eq:velocity_update}) and (\ref{eq:position_update})\;
        Evaluate the fitness of each particle\;
        Update the personal best \(p_i\) and global best \(p_g\) positions if necessary\;        
    }
\Return The global best position \(p_g\) as the final solution\;
\end{algorithm}

\subsection{PSO-GA Hybrid Algorithms}\label{sec:hybrids}

PSO and GA are both population-based optimization techniques that, while similar in their parallel processing nature, differ fundamentally in their approach to exploring and exploiting the search space. GA uses a ``competitive'' strategy, where individuals in the population compete for survival, with poorly performing individuals being replaced by offspring generated through crossover and mutation. This selection process allows GA to adaptively fine-tune the search as it evolves. In contrast, PSO operates on a ``cooperative'' principle, where particles adjust their positions based on their own best-known position and the best-known position of the swarm, without explicitly replacing any individuals. This method can lead to faster convergence in smooth, well-defined search spaces, but may struggle in more complex landscapes where directional guidance is less clear. The philosophical (together with the performance) differences between PSO and GA might be found in \cite{angeline1998evolutionary}.

It has been quickly observed that combining these two distinct approaches can yield potent hybrid algorithms, capable of addressing a broader range of complex optimization problems more effectively.
Several hybridization strategies have been proposed over the years. The overview of over 20 hybrid PSO-GA algorithms published between 2002 and 2010 can be found in \cite{thangaraj2011particle}, a good and more recent overview may be found in \cite{placzkiewicz2018hybrid,sengupta2019particle} and \cite{Shao2023PGA}. The diversity of hybridization strategies ranges from simple combinations where one algorithm initializes the population for the other, to more complex schemes where both algorithms are applied in tandem. In most cases, the integration of PSO and GA is executed either sequentially or simultaneously, where PSO typically aids in global exploration, and GA contributes to local exploitation through its crossover and mutation operations.
This study provides a comparative analysis of specific implementations of the concepts of sequential and parallel PSO-GA hybridization alongside a novel consecutive approach, focusing particularly on performance scaling with dimensionality.

\subsubsection{Sequential Approach (PGSHEA)}\label{sec:pgshea}
In sequential approaches, the two algorithms are applied one after the other in series. Such an approach, in various forms, can be easily found in the literature \cite{robinson2002particle,shi2003hybrid,juang2004hybrid,Esmin2005HPSOM,Aivaliotis2022SGA,Simaiya2024,Palaniappan2025}.

Our implementation within this approach, which we dubbed PSO-GA Sequential Hybrid Evolutionary Algorithm (PGSHEA) after \cite{shi2003hybrid}, alternates between the two optimization techniques sequentially, where the algorithm begins with one technique and switches to the other at predetermined interval. The initial population of solutions is generated using either PSO or GA (the starting algorithm is parameterized). This set of solutions is then shared between both algorithms. The PSO instance is initialized with parameters such as cognitive ($c1$), social ($c2$) coefficients, and inertia weight ($w$). The GA instance uses standard genetic operators: crossover, mutation, and selection (all three remain the same as in the standard GA implementation described in section \ref{sec:ga}). After a series of GA steps, the solutions are passed to the PSO algorithm, which initializes its population with these solutions and continues the optimization. Similarly, after a series of PSO steps, the solutions are transferred to the GA algorithm, which uses them as its starting population for further optimization.
During the switch from GA to PSO, particle velocities are typically initialized (e.g., randomly), as GA individuals do not inherently possess velocity. Conversely, when switching from PSO to GA, the velocity information associated with the particles is usually discarded.
PGSHEA continuously tracks the global best solution found so far. This solution serves as the $p_g$ for the PSO component and is preserved across switches, ensuring the search does not lose the high-quality solution, although it is not explicitly forced into the population unless selected naturally by the active algorithm's operators.
The algorithm continues to alternate between PSO and GA until the termination criterion is met. The final result is the best solution found by any particle or individual in the population after all iterations, which is returned as the output of the algorithm.

\begin{algorithm}[H]
\caption{PGSHEA}
\KwIn{Population size \textit{N}, PSO parameters (\textit{c1, c2, w}), GA parameters ($p_c$, $p_m$), starting algorithm, swap interval}
\KwOut{Best solution found}
Set current algorithm to PSO or GA based on the starting algorithm\;
Initialize population with \textit{N} individuals\;
Evaluate fitness of the initial population\;
\While{Termination criterion is not met}{
    Perform the current algorithm step on the population\;
    Update best global solution found so far\;
    \If{the swap interval is reached}{
        Switch between PSO and GA\;
    }
}
\Return The global best position as the final solution\;
\end{algorithm}


\subsubsection{Parallel Approach (PGPHEA)}
In simultaneous or parallel approaches, PSO and GA are run concurrently, and the two algorithms cooperate by sharing information or combining their results. While this approach can also be found quite widely in the literature \cite{grimaldi2004new,kao2008hybrid}, our implementation is based on the ideas from \cite{shi2003pso,shi2003hybrid} and then later presented independently by Gupta and Yadav \cite{GuptaYadav2014}.

Our implementation of this hybridization approach, which we called the PSO-GA Parallel Hybrid Evolutionary Algorithm (PGPHEA) after \cite{shi2003hybrid}, starts by dividing the population into two subpopulations: one handled by PSO and the other by GA. The subpopulations are initialized independently, with PSO generating its initial set of particles and GA generating its initial population of chromosomes.
During each iteration, both PSO and GA execute their respective steps concurrently:
PSO updates the velocity and position of each particle based on its personal best and the global best positions and
GA applies selection, crossover, and mutation to its population to create a new generation.
After both algorithms have completed their steps, the global best solution is updated by comparing the best solutions found by PSO and GA. Periodically, after a fixed number of evaluations (determined by the parametrized exchange interval), an exchange of solutions occurs between PSO and GA---the top-performing solutions from the PSO subpopulation are swapped with the top solutions from the GA subpopulation.
The exchange typically involves swapping the positional vectors of the top individuals. When solutions move from GA to the PSO subpopulation, initial velocities need to be assigned (e.g., zero or random). When solutions move from PSO to the GA subpopulation, their associated velocities are generally discarded.
The algorithm continues to run both PSO and GA in parallel until a predefined termination criterion is met.
The best solution found by either PSO or GA during the entire process is returned as the final solution.


\begin{algorithm}[H]
\caption{PGPHEA}
\KwIn{Population size \textit{N}, PSO parameters (\textit{c1, c2, w}), GA parameters ($p_c$, $p_m$), exchange interval, exchange number \textit{N\textsubscript{E}}}
\KwOut{Best solution found}
Initialize GA population \textit{P\textsubscript{GA}} with $\lceil \frac{N}{2} \rceil$ individuals and PSO population \textit{P\textsubscript{PSO}} with $N - \lceil \frac{N}{2} \rceil$ individuals\;
Evaluate fitness of both subpopulations\;
\While{Termination criterion is not met}{
    Perform PSO step on \textit{P\textsubscript{PSO}}\;
    Perform GA step on \textit{P\textsubscript{GA}}\;
    Synchronize the best global solution found so far\;
    \If{the exchange interval is reached}{
        Exchange the top \textit{N\textsubscript{E}} solutions between \textit{P\textsubscript{PSO}} and \textit{P\textsubscript{GA}}\;
    }
}
\Return The global best solution\;
\end{algorithm}

\subsubsection{Consecutive Approach (PGCHEA)}

One can think of two specific (extreme or degenerated) cases of sequential methods.
The first case is the simple two-phase hybrid scheme, where one algorithm is being used to initialize or seed the other. Although this approach is one of the oldest in evolutionary-swarm hybridizations, it has recently been successfully applied to cloud load balancing optimization and other applications \cite{robinson2002particle,Simaiya2024,Palaniappan2025}.
The other extreme is represented by consecutive approaches, where the two algorithms are applied in a highly interleaved manner---essentially, one step of one algorithm is immediately followed by a step of the other. While this tightly coupled sequential hybridization is less common, several studies have implicitly adopted its principles by closely integrating PSO and GA operations within a single iterative framework \cite{yang2007hybrid,chansamorn2022improved,Shao2023PGA}.

Building on this idea, we introduce and test a novel PSO-GA Consecutive Hybrid Evolutionary Algorithm (PGCHEA). It operates by alternately applying a PSO step and a GA step to the entire population. The algorithm initializes a population, identifies the initial global best, and then enters this alternating loop. Its core novelty lies in maintaining continuity between PSO and GA steps through dedicated information transfer mechanisms, ensuring PSO-specific information (velocity and personal best positions) is preserved and utilized across the GA steps.
This continuity is achieved via modified GA variation operators. Crossover is modified to produce offspring that inherit both velocities and personal best positions ($p_i$) from their parents, alongside their positional genetic material.
Furthermore, the concept of a personal best position is maintained for each individual regardless of the active algorithm step---a memory mechanism absent in the standard GA. This direct information inheritance distinguishes PGCHEA from a simple sequential hybrid (PGSHEA) with a swap interval of 1, where standard GA operators would typically discard velocity and personal best information, disrupting PSO's momentum.

Throughout the execution, the global best solution ($p_g$) found so far is continually updated and used to guide the search, acting as the reference point in PSO steps. The algorithm continues alternating steps until a termination criterion is met, returning the best solution found.

\begin{algorithm}[H]
\caption{PGCHEA}
\KwIn{Population size $N$, PSO parameters $(c1, c2, w)$, GA parameters $(p_c, p_m)$, starting algorithm}
\KwOut{Best solution found}
Set current algorithm to PSO or GA based on the starting algorithm\;
Initialize population $P$ with $N$ individuals\;
Evaluate fitness of the initial population\;
\While{Termination criterion is not met}{
    \If{current algorithm is PSO}{
        Perform PSO algorithm step on the population\;
        Switch to GA\;
    }
    \Else{
        Perform GA algorithm step with enhanced variation operators\;
        Update
        personal best \(p_i\) and
        global best \(p_g\) positions, if necessary\; 
        Switch to PSO\;
    }
}
\Return The global best position as the final solution\;
\end{algorithm}



\section{Experiment and result}\label{sec:experiment}

\subsection{Benchmark functions}\label{sec:functions}

In order to evaluate the performance of the proposed hybrid algorithms, a set of standard benchmark functions has been selected. The selection includes Rastrigin, Ackley, Griewank, Levy, Michalewicz, Schwefel, and Shifted Rotated Weierstrass function. Each of them presents unique characteristics, such as multiple local minima, varying degrees of complexity, and distinct search domains. The functions are tested over different dimensions (namely: \textbf{10}, \textbf{50}, \textbf{100}, \textbf{500}, and \textbf{1000}). The equations defining each function are given below, with a summary provided later in Table~\ref{tab:benchmark_functions_overview}.

\begin{small}
\begin{flalign}
f(x) &= -20 \exp\left(-0.2 \sqrt{\frac{1}{n} \sum_{i=1}^{n} x_i^2}\right) - \exp\left(\frac{1}{n} \sum_{i=1}^{n} \cos(2 \pi x_i)\right) + 20 + e && \tag{Ackley}\label{eq:ackley} \\
f(x) &= 1 + \frac{1}{4000} \sum_{i=1}^{n} x_i^2 - \prod_{i=1}^{n} \cos\left(\frac{x_i}{\sqrt{i}}\right) && \tag{Griewank}\label{eq:griewank} \\
f(x) &= \sin^2(\pi w_1) + \sum_{i=1}^{n-1} (w_i-1)^2 \left[1 + 10\sin^2(\pi w_i+1)\right] + (w_n-1)^2 \left[1 + \sin^2(2\pi w_n)\right] && \tag{Levy}\label{eq:levy} \\
f(x) &= -\sum_{i=1}^{n} \sin(x_i) \sin^{2m} \left(\frac{i x_i^2}{\pi}\right) && \tag{Michalewicz}\label{eq:michalewicz} \\
f(x) &= 10n + \sum_{i=1}^{n} \left[x_i^2 - 10 \cos(2 \pi x_i)\right] && \tag{Rastrigin}\label{eq:rastrigin} \\
f(x) &= 418.9829n - \sum_{i=1}^{n} x_i \sin(\sqrt{|x_i|}) && \tag{Schwefel}\label{eq:schwefel} \\
f(x) &= \sum_{i=1}^{n} \left(\sum_{k=0}^{20} \left[0.5^k \cos(2 \pi \cdot 3^k \cdot (x_i + 0.5))\right]\right) - n \sum_{k=0}^{20} \left[0.5^k \cos(2 \pi \cdot 3^k \cdot 0.5)\right] && \tag{Shifted Rotated Weierstrass}\label{eq:weierstrass} 
\end{flalign}
\end{small}

\begin{table}[hbt!]
\centering
\begin{threeparttable}
\caption{Benchmark functions overview}
\label{tab:benchmark_functions_overview}
{
\renewcommand{\arraystretch}{1.5}
\begin{tabularx}{\textwidth}{kkX}
\toprule
\textbf{Function Name}  & \textbf{Search Domain}\tnote{a} 
& \textbf{Fitness at Global Minimum}\tnote{a} \\
\midrule
\textbf{Ackley} & $[-32.768, 32.768]^n$ 
& $f(0) = 0$ \\
\textbf{Griewank} & $[-600, 600]^n$ 
& $f(0) = 0$ \\
\textbf{Levy} & $[-10, 10]^n$ 
& $f(1) = 0$ \\
\textbf{Michalewicz}\tnote{b} & $[0, \pi]^n$ 
& $f(x)$ is known for specific $n$ \\
\textbf{Rastrigin} & $[-5.12, 5.12]^n$ 
& $f(0) = 0$ \\
\textbf{Schwefel} & $[-500, 500]^n$ 
& $f(420.9687) \approx 0$ \\
\pbox{3cm}{\textbf{Shifted Rotated}\\ \textbf{Weierstrass}} & $[-0.5, 0.5]^n$ 
& Depends on shift and rotation\tnote{c} \\
\bottomrule
\end{tabularx}
}
\begin{tablenotes}[hang]
\item[a] $n$ -- dimensionality. $n \in \{10, 50, 100, 500, 1000\}$
\item[b] $m=10$
\item[c] The shift vector has been defined as a random vector drawn from a uniform distribution within $[-0.5, 0.5]^n$, the rotation matrix is generated as a random orthogonal matrix.
\end{tablenotes}
\end{threeparttable}
\end{table}

\subsection{Experiment setup}



The experiment was conducted to evaluate the performance of the PGSHEA, PGPHEA, and PGCHEA algorithms in comparison to the standard GA and PSO across a suite of benchmark functions, as detailed in the previous section (\ref{sec:functions}).
We adopted GA parameter values widely reported in the literature \cite{grefenstette1986optimization,Byrski_2013,hassanat2019choosing}. We then conducted a preliminary parameter tuning for the PSO and hybrid algorithms using Bayesian optimization. The resulting parameters are listed in Table \ref{tab:algorithm_parameters}.
Each algorithm was executed 10 times on each benchmark problem, with dimensionality set at \( n \in \{10, 50, 100, 500, 1000\} \). The results were averaged across runs.

The population size for all experiments was consistently set to \( N = 100 \).
The termination criterion for each algorithm was based on a maximum number of evaluations and set to 25000 for most cases and 10000 for \( n = 10 \) dimensions. This adjustment was made for clarity and practical purposes, preventing algorithms from merely converging due to the large number of evaluations, thus providing a more meaningful comparison of their performance.

\begin{table}[hbt!]
\centering
\begin{threeparttable}
\caption{Parameter settings for each algorithm}
\label{tab:algorithm_parameters}
\begin{tabular}{lccccc}
\toprule
\textbf{Parameter} & \textbf{GA} & \textbf{PSO} & \textbf{PGSHEA} & \textbf{PGPHEA} & \textbf{PGCHEA} \\
\midrule
\textbf{Crossover Rate ($\boldsymbol{p_c}$)}   & 0.9  & -    & 1.0  & 1.0  & 1.0  \\
\textbf{Mutation Rate ($\boldsymbol{p_m}$)}\tnote{a}    & $\frac{1.0}{n}$  & -    & $\frac{0.38}{n}$  & $\frac{0.37}{n}$  & $\frac{0.61}{n}$  \\
\textbf{$\boldsymbol{c_1}$}                     & -    & 1.97  & 2.63 & 0.01 & 1.85 \\
\textbf{$\boldsymbol{c_2}$}                     & -    & 0.94  & 0.21 & 0.26 & 0.5 \\
\textbf{$\boldsymbol{w}$}             & -    & 0.56  & 0.01 & 0.17 & 1.53 \\
\textbf{Exchange Interval}         & -    & -     & 13   & 13   & -    \\
\textbf{Exchange Number}           & -    & -     & -    & 7    & -    \\
\textbf{Starting Algorithm}        & -    & -     & PSO  & -    & PSO  \\
\bottomrule
\end{tabular}
\begin{tablenotes}[hang]
\item[a] $n$ -- dimensionality. $n \in \{10, 50, 100, 500, 1000\}$
\end{tablenotes}
\end{threeparttable}
\end{table}

\subsection{Experiment results}

\begin{figure}[hbt!]
    \centering
    \begin{subfigure}[b]{0.48\textwidth}
        \centering
        \includegraphics[width=0.99\textwidth]{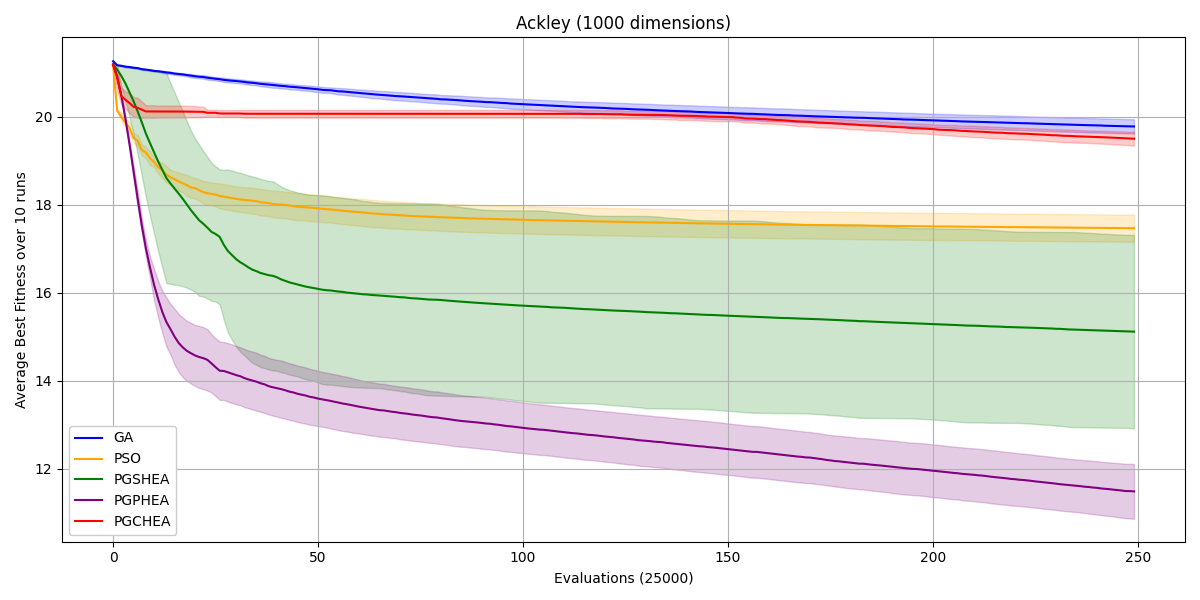}
        \caption{}
    \end{subfigure}
    \begin{subfigure}[b]{0.48\textwidth}
        \centering
        \includegraphics[width=0.99\textwidth]{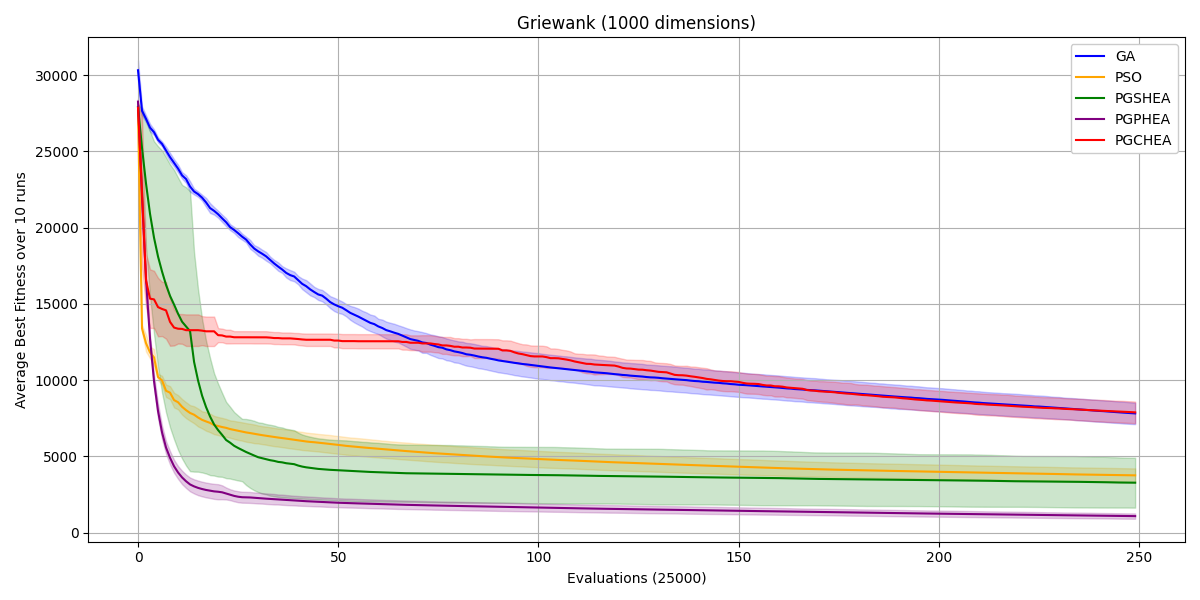}
        \caption{}
    \end{subfigure}
    \begin{subfigure}[b]{0.48\textwidth}
        \centering
        \includegraphics[width=0.99\textwidth]{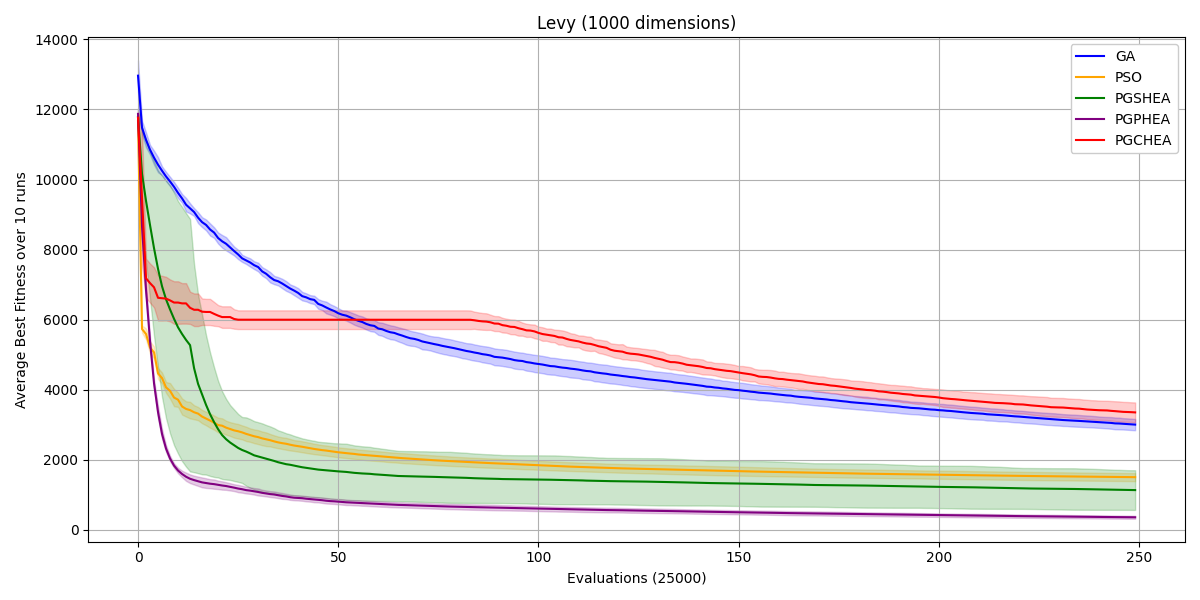}
        \caption{}
    \end{subfigure}
    \begin{subfigure}[b]{0.48\textwidth}
        \centering
        \includegraphics[width=0.99\textwidth]{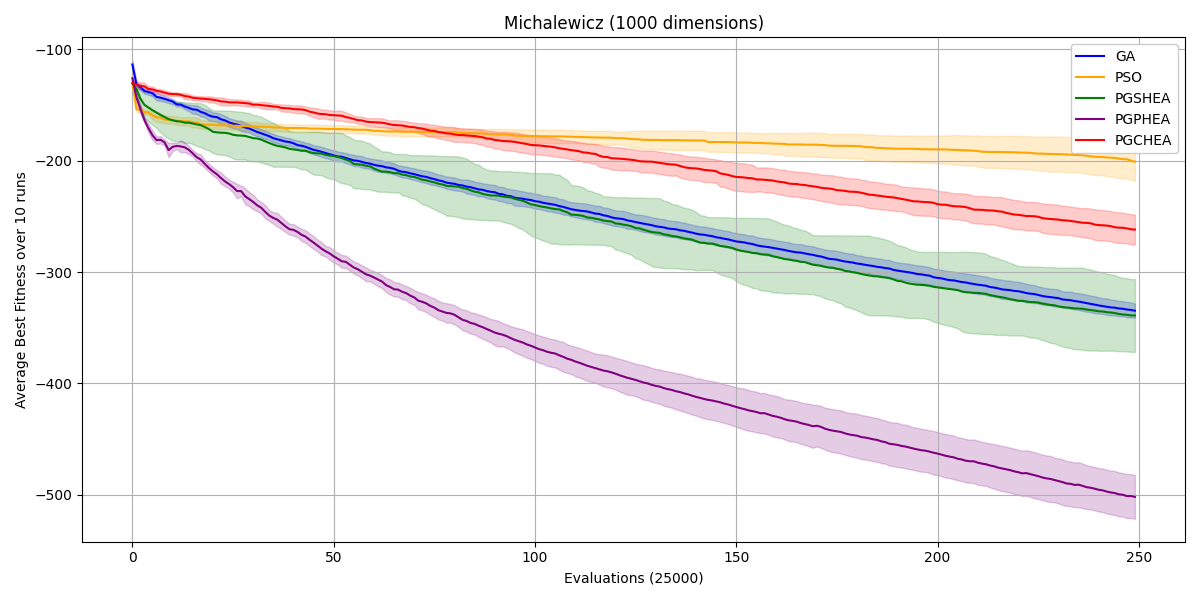}
        \caption{}
    \end{subfigure}
    \begin{subfigure}[b]{0.48\textwidth}
        \centering
        \includegraphics[width=0.99\textwidth]{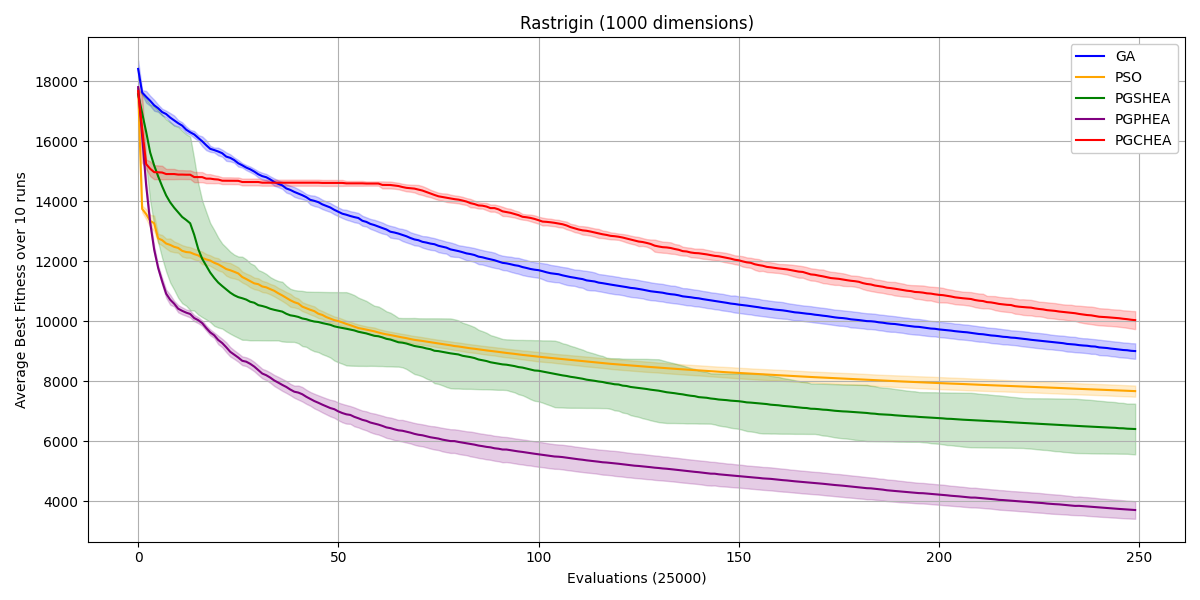}
        \caption{}
    \end{subfigure}
    \begin{subfigure}[b]{0.48\textwidth}
        \centering
        \includegraphics[width=0.99\textwidth]{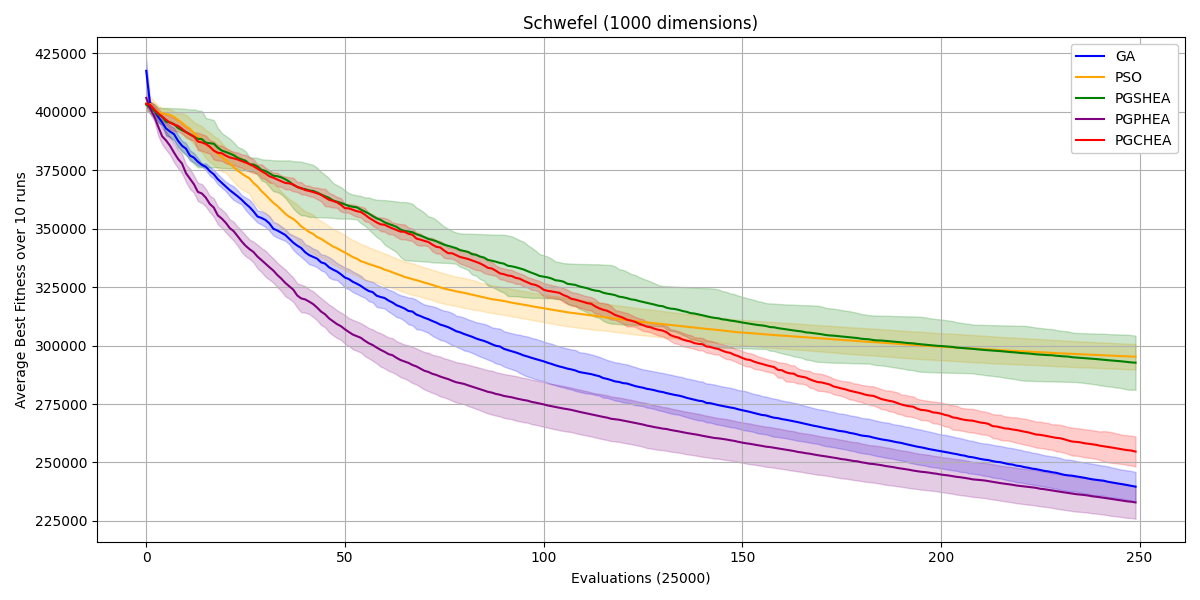}
        \caption{}
    \end{subfigure}
    \caption{Convergence analysis of GA (\textcolor{blue}{blue}), PSO (\textcolor{Goldenrod}{yellow}), PGSHEA (\textcolor{OliveGreen}{green}), PGPHEA (\textcolor{RedViolet}{magenta}), and PGCHEA (\textcolor{red}{red}) on six benchmark functions with 1000 dimensions, evaluated over 25,000 iterations: (a) Ackley, (b) Griewank, (c) Levy, (d) Michalewicz, (e) Rastrigin, and (f) Schwefel.}
    \label{fig:other}
\end{figure}

\begin{figure}[hbt!]
    \centering
    \begin{subfigure}[b]{0.48\textwidth}
        \centering
        \includegraphics[width=0.99\textwidth]{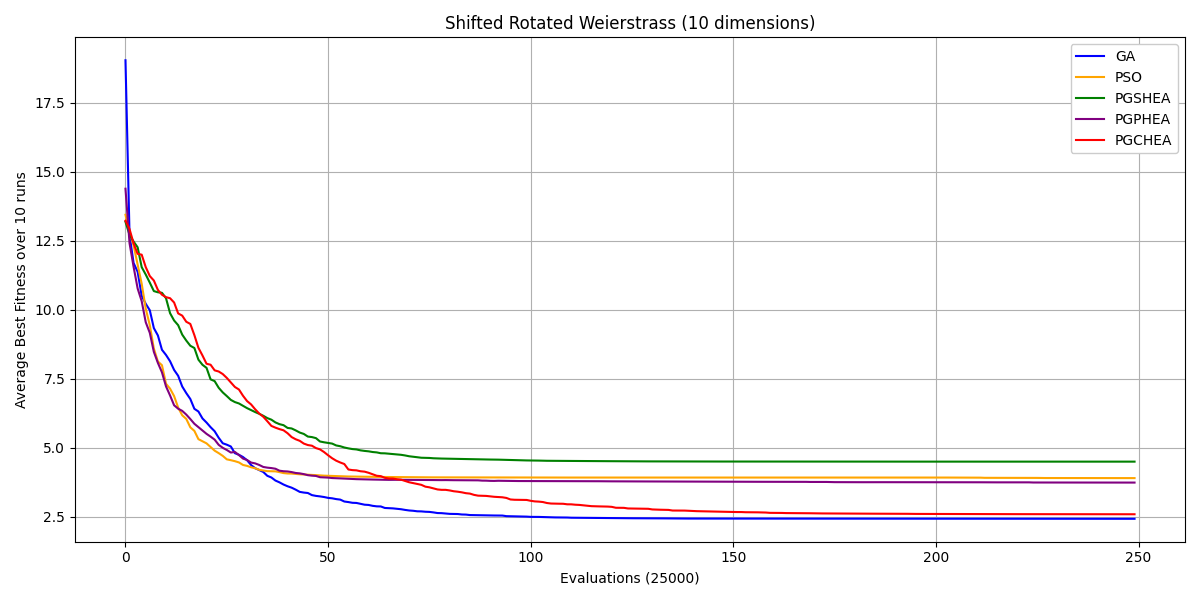}
    \end{subfigure}
    \begin{subfigure}[b]{0.48\textwidth}
        \centering
        \includegraphics[width=0.99\textwidth]{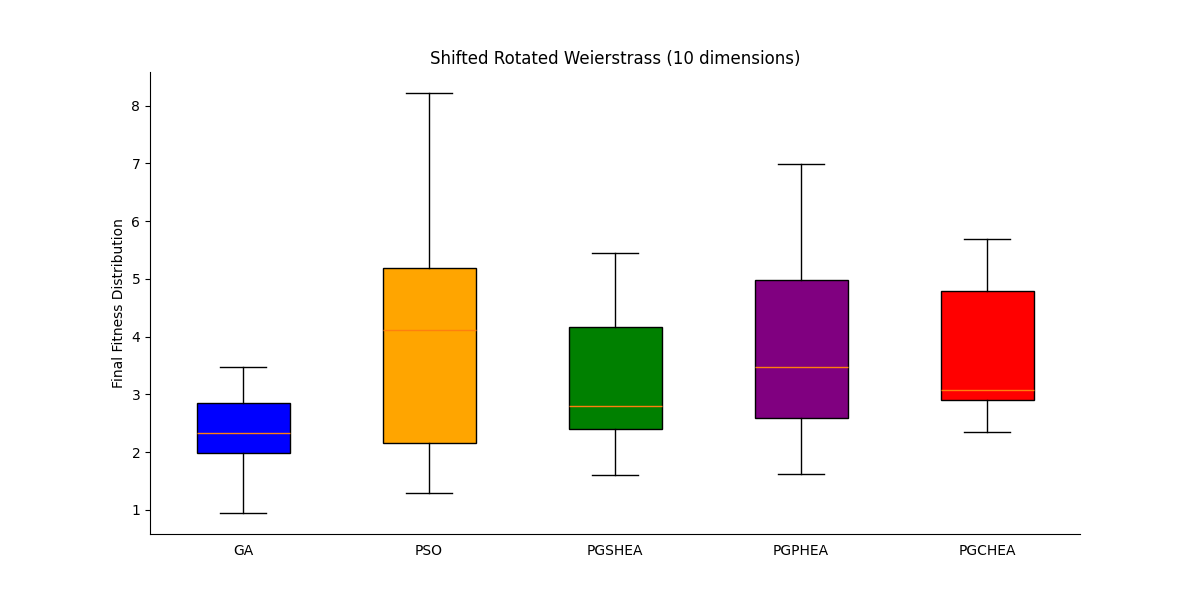}
    \end{subfigure}
    \begin{subfigure}[b]{0.48\textwidth}
        \centering
        \includegraphics[width=0.99\textwidth]{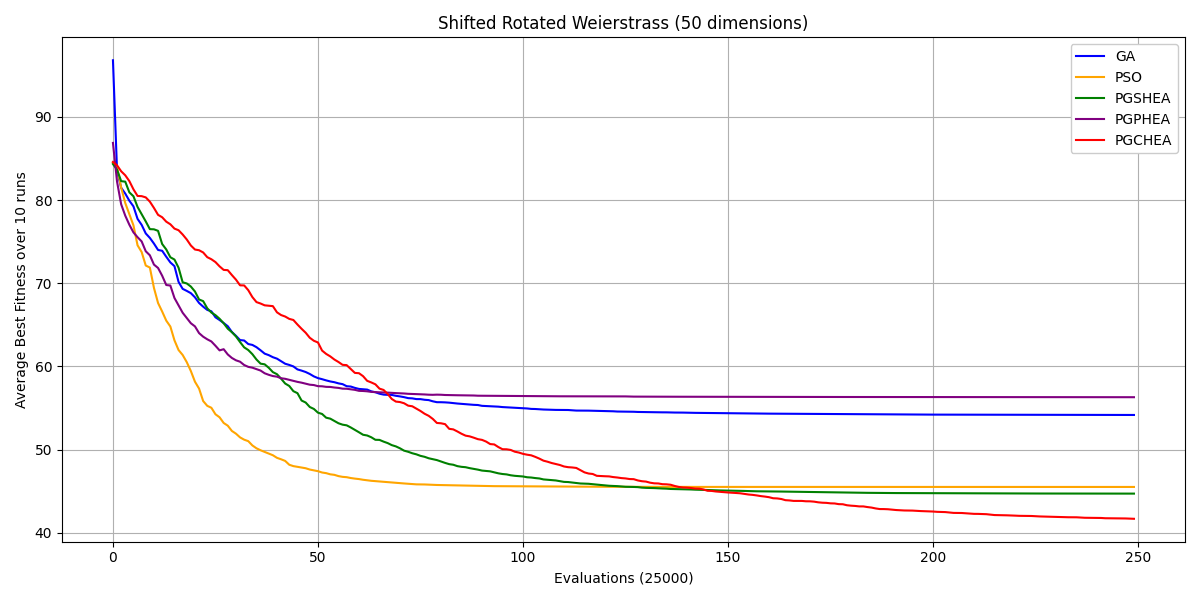}
    \end{subfigure}
    \begin{subfigure}[b]{0.48\textwidth}
        \centering
        \includegraphics[width=0.99\textwidth]{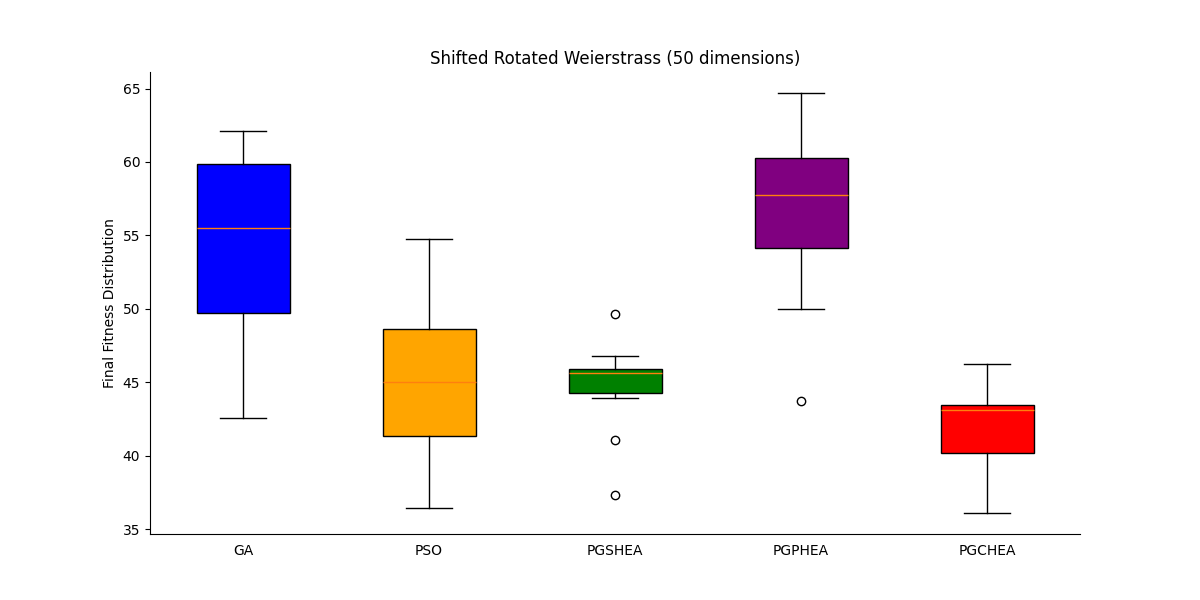}
    \end{subfigure}
    \begin{subfigure}[b]{0.48\textwidth}
        \centering
        \includegraphics[width=0.99\textwidth]{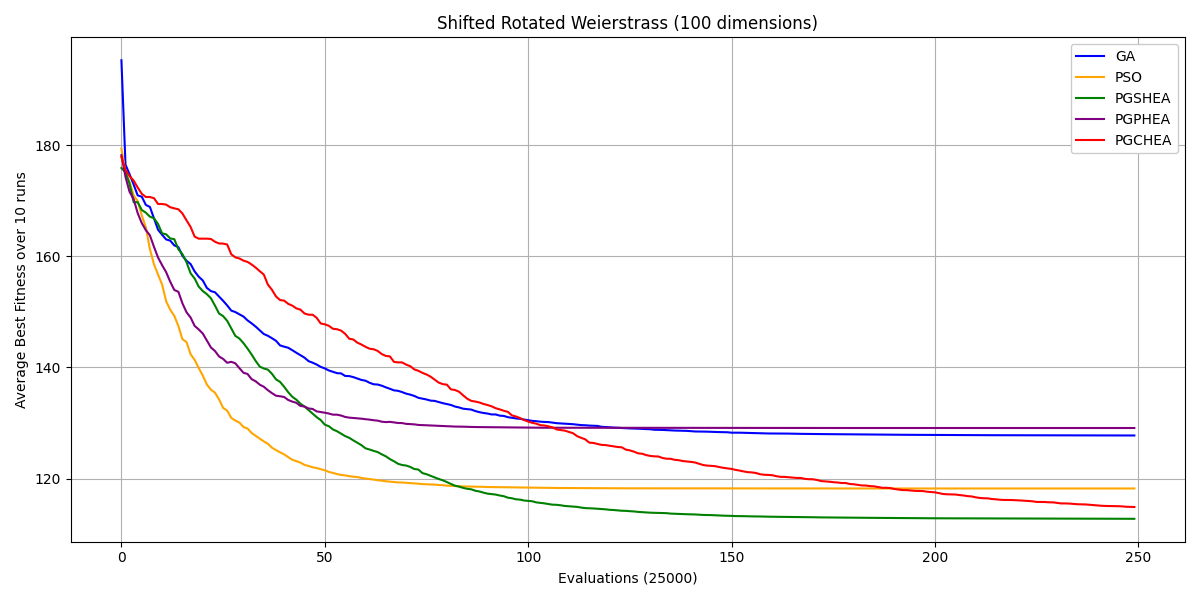}
    \end{subfigure}
    \begin{subfigure}[b]{0.48\textwidth}
        \centering
        \includegraphics[width=0.99\textwidth]{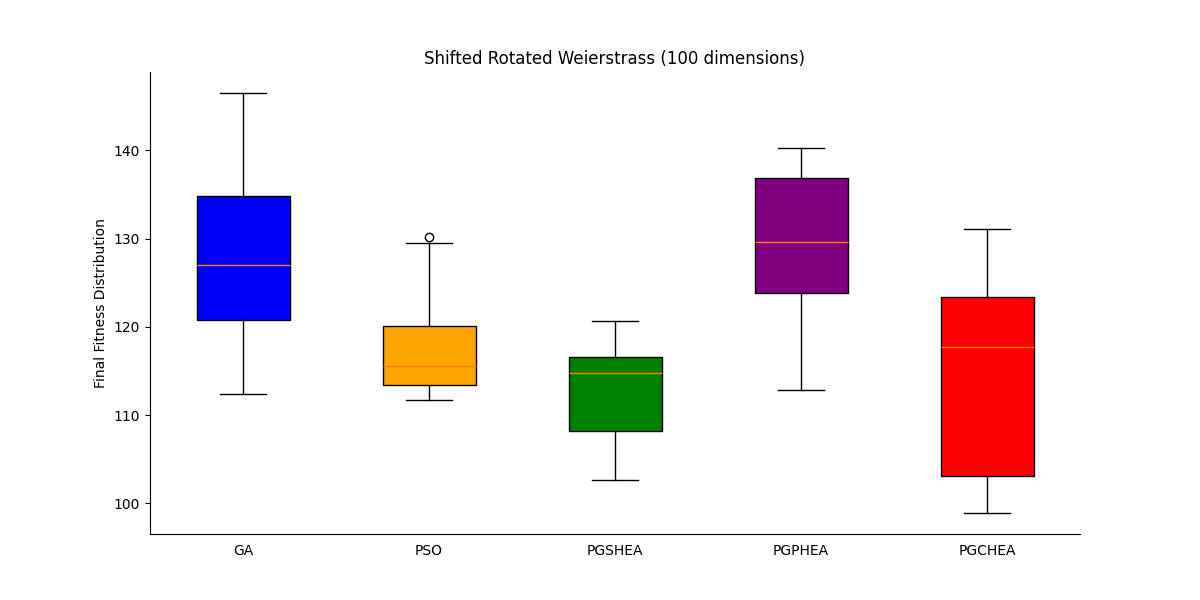}
    \end{subfigure}
    \begin{subfigure}[b]{0.48\textwidth}
        \centering
        \includegraphics[width=0.99\textwidth]{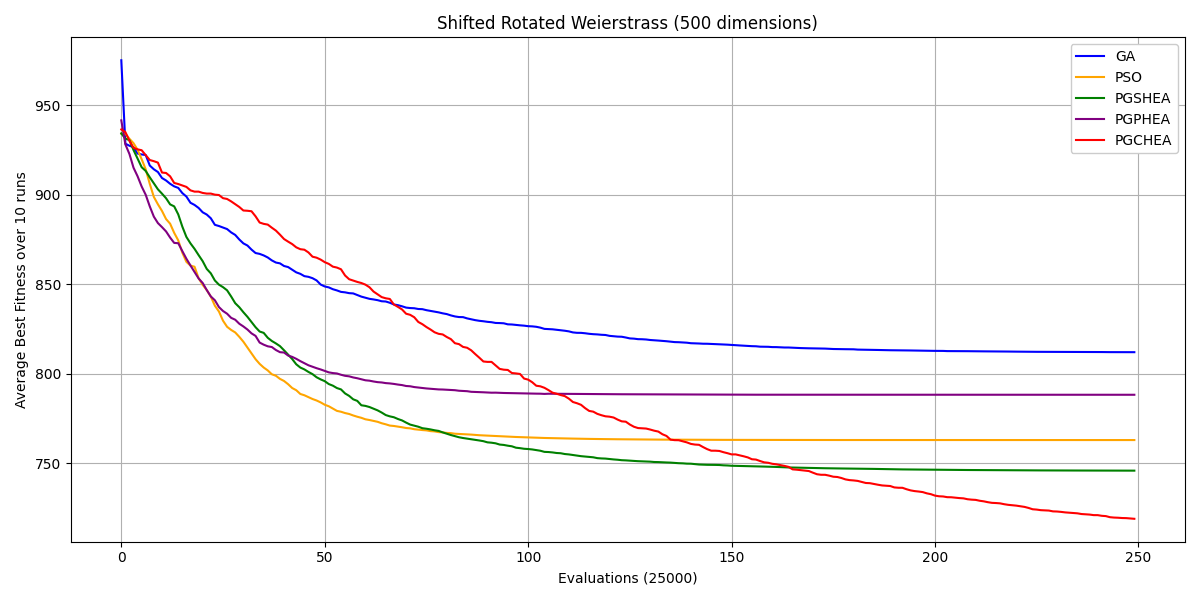}
    \end{subfigure}
    \begin{subfigure}[b]{0.48\textwidth}
        \centering
        \includegraphics[width=0.99\textwidth]{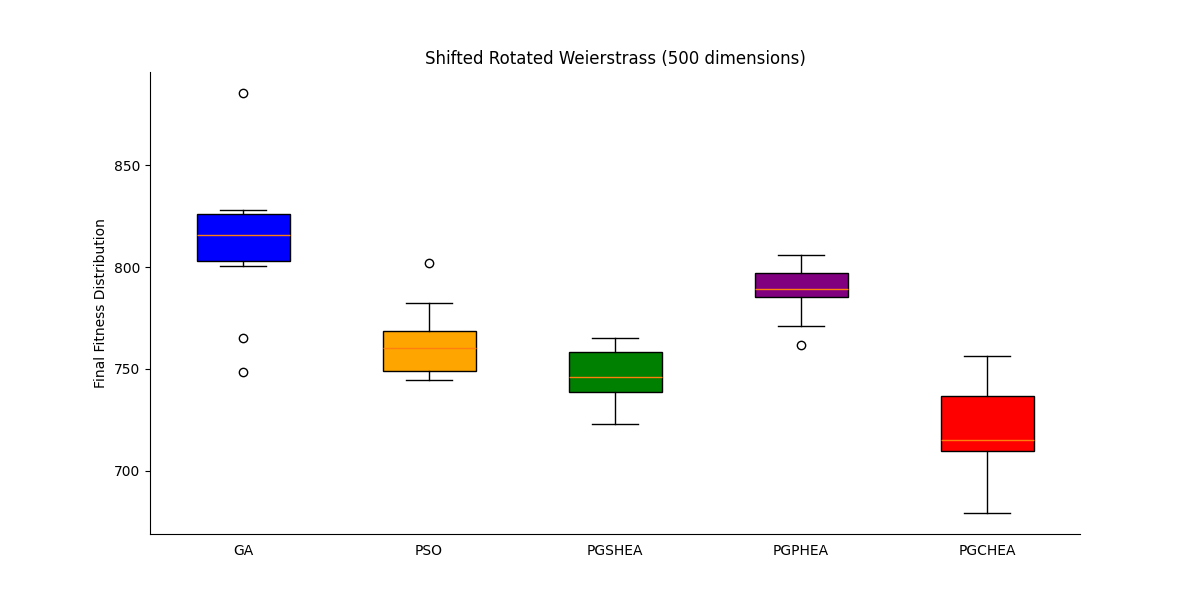}
    \end{subfigure}
    \begin{subfigure}[b]{0.48\textwidth}
        \centering
        \includegraphics[width=0.99\textwidth]{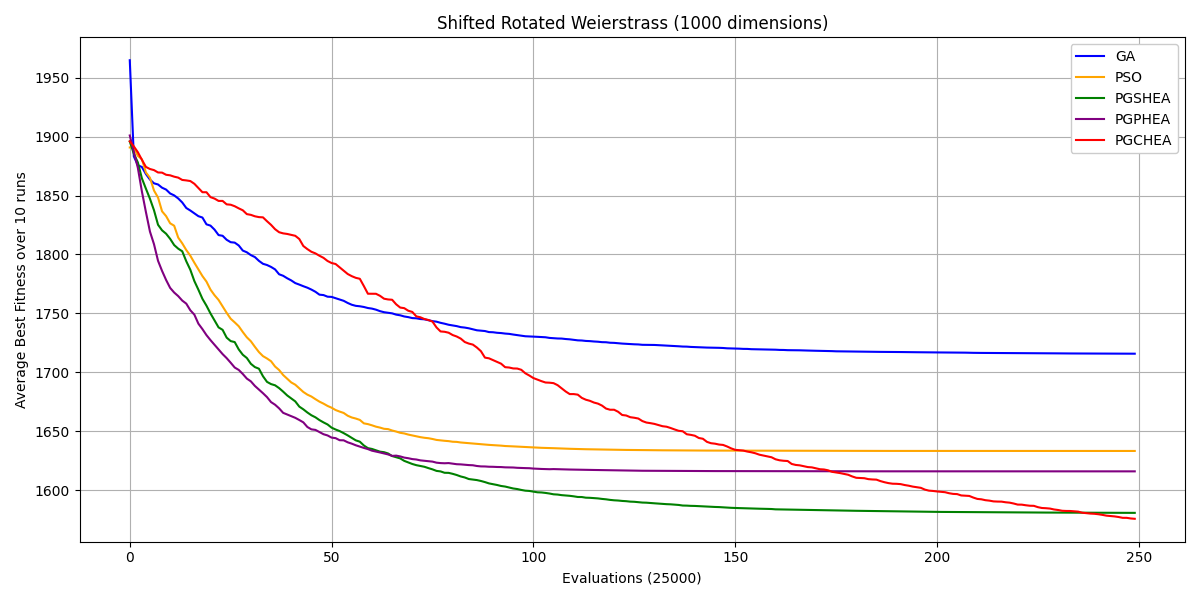}
    \end{subfigure}
    \begin{subfigure}[b]{0.48\textwidth}
        \centering
        \includegraphics[width=0.99\textwidth]{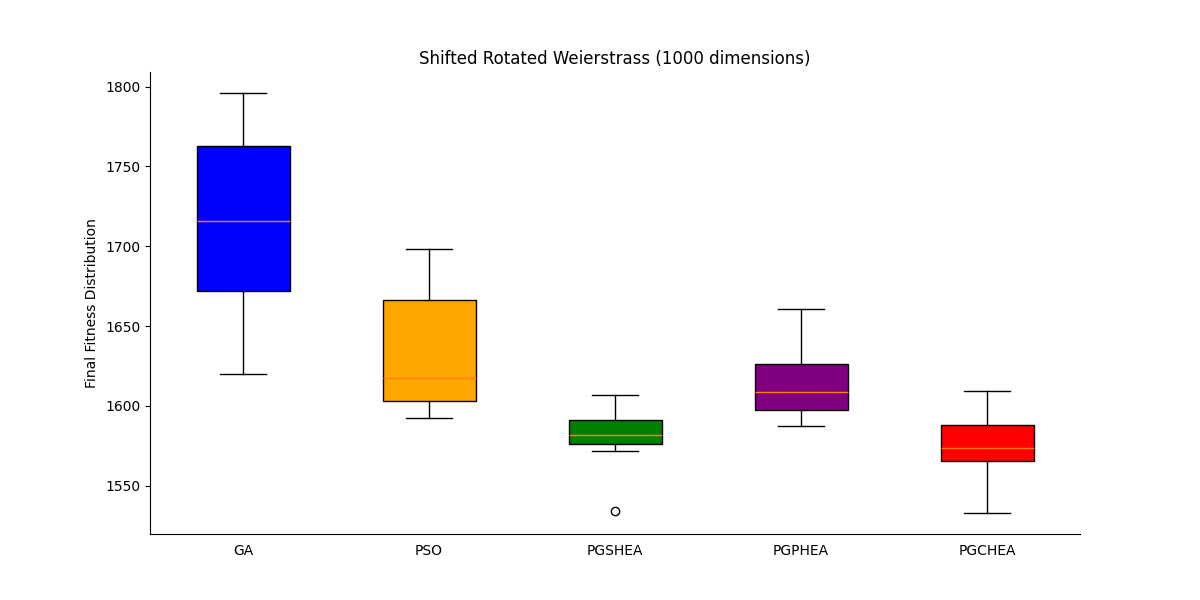}
    \end{subfigure}
    \caption{Convergence plots (left) and  final fitness distribution analysis (right) 
    of GA (\textcolor{blue}{blue}), PSO (\textcolor{Goldenrod}{yellow}), PGSHEA (\textcolor{OliveGreen}{green}), PGPHEA (\textcolor{RedViolet}{magenta}), and PGCHEA (\textcolor{red}{red}) on the Shifted Rotated Weierstrass function for 10, 50, 100, 500, and 1000 dimensions respectively (in top-down order).}
    \label{fig:srw}
\end{figure}

Table \ref{tab:experiment_results} presents the comparative performance results of GA, PSO, PGSHEA, PGPHEA, and PGCHEA across various benchmark functions. The table shows the average fitness obtained by each algorithm for each problem after 25000 evaluations (for most dimensions) and 10000 evaluations (for the smallest dimensional cases). The best-performing algorithm for each problem and dimension is highlighted in bold. Detailed results, including convergence plots and performance analysis, can be found in the Figures~\ref{fig:other} and~\ref{fig:srw}.

\begin{table}[hbt!]
\centering
\caption{Experiment results}
\label{tab:experiment_results}
\resizebox{\textwidth}{!}{
\begin{tabular}{|c|c|c|c|c|c|c|c|}
\hline
\textbf{Problem} & \textbf{Dim. ($\boldsymbol{n}$)} & \textbf{Eval.} & \textbf{GA} & \textbf{PSO} & \textbf{PGSHEA} & \textbf{PGPHEA} & \textbf{PGCHEA} \\
\hline
& 10 & 10000 & 0.0812 & \textbf{0.0000} & 0.0942 & 0.0185 & 0.2991 \\ 
& 50 & 25000 & 0.4264 & 4.6518 & 1.6092 & \textbf{0.1215} & 2.3408 \\ 
\textbf{Ackley} & 100 & 25000 & 3.2003 & 9.1545 & 5.0698 & \textbf{1.8037} & 6.3891 \\ 
& 500 & 25000 & 18.3729 & 16.5278 & 13.8260 & \textbf{8.9708} & 17.6277 \\ 
& 1000 & 25000 & 19.7747 & 17.4644 & 15.1160 & \textbf{11.4876} & 19.4970 \\ 
\hline
& 10 & 10000 & 0.1537 & \textbf{0.0654} & 0.1563 & 0.0807 & 0.4173 \\ 
& 50 & 25000 & 1.0387 & \textbf{0.0294} & 1.2411 & 0.7355 & 1.3394 \\ 
\textbf{Griewank} & 100 & 25000 & 2.4290 & 2.7071 & 10.5749 & \textbf{1.3514} & 13.1571 \\ 
& 500 & 25000 & 1877.3427 & 946.6615 & 1085.4084 & \textbf{213.7523} & 2267.2059 \\ 
& 1000 & 25000 & 7815.1528 & 3752.3912 & 3271.2027 & \textbf{1078.5572} & 7882.8616 \\ 
\hline
& 10 & 10000 & 0.0003 & \textbf{0.0000} & 0.0006 & \textbf{0.0000} & 0.0019 \\ 
& 50 & 25000 & 0.3300 & 6.6185 & \textbf{0.2859} & 0.3596 & 0.6722 \\ 
\textbf{Levy} & 100 & 25000 & 11.0033 & 33.9640 & 12.5885 & \textbf{0.7897} & 26.2848 \\ 
& 500 & 25000 & 756.2910 & 436.1920 & 318.3160 & \textbf{85.6619} & 933.4690 \\ 
& 1000 & 25000 & 3006.4498 & 1506.2417 & 1137.4485 & \textbf{361.4372} & 3354.4831 \\ 
\hline
& 10 & 10000 & \textbf{-9.4219} & -8.8826 & -9.3585 & -9.3291 & -9.3618 \\ 
& 50 & 25000 & -41.2735 & -35.7822 & -41.6125 & \textbf{-42.9070} & -40.9484 \\ 
\textbf{Michalewicz} & 100 & 25000 & -72.6364 & -59.8520 & -74.4890 & \textbf{-76.2109} & -69.3091 \\ 
& 500 & 25000 & -224.7013 & -132.8061 & -211.3493 & \textbf{-336.4667} & -175.3285 \\ 
& 1000 & 25000 & -334.7268 & -200.9745 & -339.0732 & \textbf{-501.9018} & -261.8926 \\ 
\hline
& 10 & 10000 & \textbf{0.1895} & 5.4139 & 2.2492 & 0.4095 & 2.6509 \\ 
& 50 & 25000 & 28.0475 & 85.3689 & 41.7002 & \textbf{14.0602} & 50.8819 \\ 
\textbf{Rastrigin}  & 100 & 25000 & 164.9228 & 265.5786 & 148.5575 & \textbf{48.3258} & 221.7512 \\ 
& 500 & 25000 & 3143.9534 & 3015.7626 & 2291.8820 & \textbf{984.8022} & 3665.6620 \\ 
& 1000 & 25000 & 9010.0454 & 7674.9698 & 6411.1468 & \textbf{3710.6077} & 10041.3225 \\ 
\hline
& 10 & 10000 & 94.9378 & 628.0398 & 166.1491 & 213.2274 & \textbf{60.1219} \\ 
& 50 & 25000 & \textbf{1953.1047} & 9153.6703 & 3646.7280 & 2320.6635 & 2186.8971 \\ 
\textbf{Schwefel} & 100 & 25000 & \textbf{7120.7331} & 19838.0034 & 12580.6665 & 8033.4714 & 8700.4964 \\ 
& 500 & 25000 & 91144.4005 & 128370.9587 & 121689.7331 & \textbf{89486.0379} & 103279.6629 \\ 
& 1000 & 25000 & 239641.4942 & 295232.4449 & 292685.6819 & \textbf{232898.2801} & 254663.7326 \\ 
\hline
& 10 & 10000 & \textbf{2.3408} & 4.1009 & 3.2057 & 3.8153 & 3.6618 \\ 
\textbf{Shifted} & 50 & 25000 & 54.1631 & 45.5086 & 44.7081 & 56.2984 & \textbf{41.6800} \\ 
\textbf{Rotated} & 100 & 25000 & 127.7596 & 118.2235 & \textbf{112.7674} & 129.1044 & 114.8938 \\ 
\textbf{Weierstrass} & 500 & 25000 & 812.1367 & 763.0698 & 745.9866 & 788.3598 & \textbf{719.0998} \\ 
& 1000 & 25000 & 1715.7651 & 1633.2264 & 1580.6910 & 1615.9141 & \textbf{1575.6579} \\ 
\hline
\end{tabular}
}
\end{table}

\begin{table}[hbt!]
\centering
\caption{Best performing algorithms per problem and dimension}
\label{tab:best_algorithms}
\resizebox{\textwidth}{!}{
\begin{tabular}{|l|c|c|c|c|c|}
\hline
\textbf{Problem/Dimension} & \textbf{10} & \textbf{50} & \textbf{100} & \textbf{500} & \textbf{1000} \\
\hline
\textbf{Ackley} & \textcolor{orange}{\textbf{PSO}} & \textcolor{purple}{\textbf{PGPHEA}} & \textcolor{purple}{\textbf{PGPHEA}} & \textcolor{purple}{\textbf{PGPHEA}} & \textcolor{purple}{\textbf{PGPHEA}} \\
\hline
\textbf{Griewank} & \textcolor{orange}{\textbf{PSO}} & \textcolor{orange}{\textbf{PSO}} & \textcolor{purple}{\textbf{PGPHEA}} & \textcolor{purple}{\textbf{PGPHEA}} & \textcolor{purple}{\textbf{PGPHEA}} \\
\hline
\textbf{Levy} & \textcolor{orange}{\textbf{PSO}}/\textcolor{purple}{\textbf{PGPHEA}} & \textcolor[rgb]{0.0, 0.5, 0.0}{\textbf{PGSHEA}} & \textcolor{purple}{\textbf{PGPHEA}} & \textcolor{purple}{\textbf{PGPHEA}} & \textcolor{purple}{\textbf{PGPHEA}} \\
\hline
\textbf{Michalewicz} & \textcolor{blue}{\textbf{GA}} & \textcolor{purple}{\textbf{PGPHEA}} & \textcolor{purple}{\textbf{PGPHEA}} & \textcolor{purple}{\textbf{PGPHEA}} & \textcolor{purple}{\textbf{PGPHEA}} \\
\hline
\textbf{Rastrigin} & \textcolor{blue}{\textbf{GA}} & \textcolor{purple}{\textbf{PGPHEA}} & \textcolor{purple}{\textbf{PGPHEA}} & \textcolor{purple}{\textbf{PGPHEA}} & \textcolor{purple}{\textbf{PGPHEA}} \\
\hline
\textbf{Schwefel} & \textcolor{red}{\textbf{PGCHEA}} & \textcolor{blue}{\textbf{GA}} & \textcolor{blue}{\textbf{GA}} & \textcolor{purple}{\textbf{PGPHEA}} & \textcolor{purple}{\textbf{PGPHEA}} \\
\hline
\textbf{Shifted Rotated Weierstrass} & \textcolor{blue}{\textbf{GA}} & \textcolor{red}{\textbf{PGCHEA}} & \textcolor[rgb]{0.0, 0.5, 0.0}{\textbf{PGSHEA}} & \textcolor{red}{\textbf{PGCHEA}} & \textcolor{red}{\textbf{PGCHEA}} \\
\hline
\end{tabular}
}
\end{table}

We have systematically performed various statistical
testing on the quantitative results we have
obtained. First, we have applied the Shapiro‐Wilk test
with a significance threshold of 0.05 to assess whether the observed samples followed a normal distribution. The outcomes were mixed, as the null hypothesis was rejected for some samples while it could not be rejected for others. Consequently, we employed the non-parametric Kruskal-Wallis test to determine if the cumulative distribution functions differed among the groups. The results of the test were statistically significant. This was followed by pairwise comparisons using
Dunn’s test to identify which pairs exhibited statistically significant differences. Table~\ref{tab:non_significant_pairs} presents the algorithm pairs that did not show statistically significant differences
(assuming the above-mentioned significance level $\alpha$)
when compared against the best-performing algorithm for higher-dimensional problems.

\begin{footnotesize}
\begin{table}[hbt!]
\centering
\caption{Non-significant algorithm pairs with p-values}
\begin{tabular}{llcc}
\toprule
\textbf{Problem} & \textbf{Dimension} & \textbf{Algorithm Pair} & \textbf{p-value} \\
\midrule
Ackley & 100 & GA vs PGPHEA & 0.1084 \\
Griewank & 100 & PSO vs PGPHEA & 0.0666 \\
Michalewicz & 100 & GA vs PGPHEA & 0.1761 \\
Michalewicz & 100 & PGSHEA vs PGPHEA & 0.4625 \\
Schwefel & 100 & GA vs PGPHEA & 0.2442 \\
Schwefel & 100 & GA vs PGCHEA & 0.0906 \\
Shifted Rotated Weierstrass & 100 & PSO vs PGSHEA & 0.4769 \\
Shifted Rotated Weierstrass & 100 & PGSHEA vs PGCHEA & 0.6247 \\
Rastrigin & 500 & PGSHEA vs PGPHEA & 0.1155 \\
Ackley & 500 & PGSHEA vs PGPHEA & 0.0577 \\
Levy & 500 & PGSHEA vs PGPHEA & 0.0577 \\
Michalewicz & 500 & GA vs PGPHEA & 0.0577 \\
Schwefel & 500 & GA vs PGPHEA & 0.6676 \\
Shifted Rotated Weierstrass & 500 & PGSHEA vs PGCHEA & 0.2152 \\
Rastrigin & 1000 & PGSHEA vs PGPHEA & 0.1389 \\
Ackley & 1000 & PGSHEA vs PGPHEA & 0.1038 \\
Schwefel & 1000 & GA vs PGPHEA & 0.4337 \\
Shifted Rotated Weierstrass & 1000 & PGSHEA vs PGCHEA & 0.7590 \\
\bottomrule
\end{tabular}
\label{tab:non_significant_pairs}
\end{table}
\end{footnotesize}

\section{Conclusions}

The hybrid algorithms (PGSHEA, PGPHEA, and PGCHEA) generally outperform the standard GA and PSO in most cases, especially as the dimensionality of the problems increases. This suggests that combining the strengths of both GA and PSO within these hybrid frameworks provides a more robust approach to optimization, particularly for complex and high-dimensional problems.
Also, the performance gap between the algorithms becomes more pronounced as the number of dimensions increases. For instance, in lower-dimensional problems (e.g., 10 dimensions), GA and PSO sometimes achieve competitive results. However, in higher-dimensional problems (e.g., 500 and 1000 dimensions), the hybrid algorithms, particularly PGPHEA, often show superior performance, indicating that the hybrid approaches are better suited to handle the complexity associated with larger search spaces.

The standard evolutionary algorithms perform well in lower dimensions (at least on certain functions), but generally struggle as the problem complexity increases. Particularly, PSO performs well on problems like Levy and Griewank in lower dimensions, where its ability to explore the search space leads to good initial results. However, as indicated by the plots in the appendix, the same exploration tendency usually leads to premature convergence to local minima.

The consistent performance of PGPHEA across a wide range of problems and dimensions suggests that it is the most versatile and robust technique among all tested algorithms. It adapts well to different problem types, making it a strong candidate for general-purpose optimization tasks.

Notably, the novel PGCHEA algorithm, while not the best performer overall, demonstrated particular strength on the complex Shifted Rotated Weierstrass function, highlighting that its unique mechanism for continuous information transfer can be beneficial for certain challenging landscapes with intricate, non-separable, or deceptive structures and warrants further investigation.
This result reinforces the ``No Free Lunch Theorem'' in optimization, which states that no single algorithm can outperform others across all problem types,
highlighting the inherent complexity of optimization and underscores the limitations of relying on a single algorithm to excel across diverse problem landscapes.

\begin{credits}
\subsubsection{\ackname} The research presented in this paper has been financially supported by: Polish National Science Center Grant no. 2019/35/O/ST6/00570 ``Socio-cognitive inspirations in classic metaheuristics'';  Polish Ministry of Science and Higher Education funds assigned to AGH University of Science and Technology, program „Excellence initiative – research university" for the AGH University of Krakow. ARTIQ project – Polish National Science Center:DEC-2021/01/2/ST6/00004, Polish National Center for Research and Development, DWP/ARTIQI/426/2023 (MKD)

\subsubsection{\discintname}
The authors have no competing interests to declare that are
relevant to the content of this article. 
\end{credits}

\nocite{DBLP:journals/jocs/TurekSKAPBK16,Byrski_Swiderska_Lasisz_Kisiel-Dorohinicki_Lenaerts_Samson_Indurkhya_2018,DBLP:journals/csse/ByrskiDK12}

\bibliographystyle{splncs04}
\bibliography{pgh}

\begin{thebibliography}{10}
\providecommand{\url}[1]{\texttt{#1}}
\providecommand{\urlprefix}{URL }
\providecommand{\doi}[1]{https://doi.org/#1}

\bibitem{Aivaliotis2022SGA}
Aivaliotis-Apostolopoulos, P., Loukidis, D.: Swarming genetic algorithm: A
  nested fully coupled hybrid of genetic algorithm and particle swarm
  optimization. PLOS ONE  \textbf{17}(9),  e0275094 (2022).
  \doi{10.1371/journal.pone.0275094}

\bibitem{angeline1998evolutionary}
Angeline, P.J.: Evolutionary optimization versus particle swarm optimization:
  Philosophy and performance differences. In: Proceedings of the Seventh Annual
  Conference on Evolutionary Programming. pp. 601--610. Springer, Berlin,
  Heidelberg (1998). \doi{10.1007/BFb0040811}

\bibitem{Byrski_2013}
Byrski, A.: Tuning of agent-based computing. Computer Science  \textbf{14}(3),
  491--512 (2013). \doi{10.7494/csci.2013.14.3.491}

\bibitem{DBLP:journals/csse/ByrskiDK12}
Byrski, A., Debski, R., Kisiel{-}Dorohinicki, M.: Agent-based computing in an
  augmented cloud environment. Comput. Syst. Sci. Eng.  \textbf{27}(1) (2012)

\bibitem{Byrski_Swiderska_Lasisz_Kisiel-Dorohinicki_Lenaerts_Samson_Indurkhya_2018}
Byrski, A., Swiderska, E., Lasisz, J., Kisiel-Dorohinicki, M., Lenaerts, T.,
  Samson, D., Indurkhya, B.: Emergence of population structure in
  socio-cognitively inspired ant colony optimization. Computer Science
  \textbf{19}(1),  81--98 (2018). \doi{10.7494/csci.2018.19.1.2594}

\bibitem{chansamorn2022improved}
Chansamorn, S., Somgiat, W.: Improved particle swarm optimization using
  evolutionary algorithm. In: 2022 19th International Joint Conference on
  Computer Science and Software Engineering ({JCSSE}). pp. 319--324. IEEE
  (2022). \doi{10.1109/JCSSE54890.2022.9836238}

\bibitem{eberhart1995new}
Eberhart, R.C., Kennedy, J.: A new optimizer using particle swarm theory. In:
  Proceedings of the Sixth International Symposium on Micro Machine and Human
  Science. pp. 39--43. IEEE (1995). \doi{10.1109/MHS.1995.494215}

\bibitem{Esmin2005HPSOM}
Esmin, A.A.A., Lambert-Torres, G., Zambroni~de Souza, A.C.: A hybrid {Particle}
  {Swarm} {Optimization} applied to loss power minimization. IEEE Transactions
  on Power Systems  \textbf{20}(2),  859--866 (2005).
  \doi{10.1109/TPWRS.2005.846049}

\bibitem{goldberg1989genetic}
Goldberg, D.E.: Genetic Algorithms in Search, Optimization, and Machine
  Learning. Addison-Wesley, Reading, MA (1989)

\bibitem{grefenstette1986optimization}
Grefenstette, J.J.: Optimization of control parameters for genetic algorithms.
  IEEE Transactions on Systems, Man, and Cybernetics  \textbf{16}(1),  122--128
  (1986)

\bibitem{grimaldi2004new}
Grimaldi, E., Grimaccia, F., Mussetta, M., Pirinoli, P., Zich, R.: A new hybrid
  genetical-swarm algorithm for electromagnetic optimization. In: ICCEA 2004
  Proceedings. 3rd International Conference on Computational Electromagnetics
  and Its Applications. pp. 157--160 (2004)

\bibitem{GuptaYadav2014}
Gupta, M., Yadav, R.: New improved fractional order differentiator models based
  on optimized digital differentiators. The Scientific World Journal  (2014)

\bibitem{hassanat2019choosing}
Hassanat, A.B., Almohammadi, K., Alkafaween, E., Abunawas, E., Hammouri, A.,
  Prasath, V.B.S.: Choosing mutation and crossover ratios for genetic
  algorithms—a~review with a new dynamic approach. Information
  \textbf{10}(12), ~390 (2019)

\bibitem{holland1975adaptation}
Holland, J.H.: Adaptation in Natural and Artificial Systems. University of
  Michigan Press, Ann Arbor, MI, 1st edn. (1975)

\bibitem{juang2004hybrid}
Juang, C.M.: A hybrid of genetic algorithm and particle swarm optimization for
  recurrent network design. IEEE Transactions on Systems, Man, and Cybernetics,
  Part B (Cybernetics)  \textbf{34}(2),  997--1006 (2004)

\bibitem{kao2008hybrid}
Kao, Y.C., Zahara, E.: A hybrid genetic algorithm and particle swarm
  optimization for multimodal functions. Applied Soft Computing  \textbf{8}(2),
   849--857 (2008)

\bibitem{kennedy1995particle}
Kennedy, J., Eberhart, R.C.: Particle swarm optimization. In: Proceedings of
  the IEEE International Conference on Neural Networks. pp. 1942--1948. IEEE
  (1995)

\bibitem{Palaniappan2025}
Palaniappan, S.C., Ponnuswamy, P.P.: Task offloading in edge computing using
  integrated particle swarm optimization and genetic algorithm. Advances in
  Science and Technology Research Journal  \textbf{19}(1),  371--380 (2025)

\bibitem{placzkiewicz2018hybrid}
Placzkiewicz, L., Sendera, M., Szlachta, A., Paciorek, M., Byrski, A.,
  Kisiel-Dorohinicki, M., Godzik, M.: Hybrid swarm and agent-based evolutionary
  optimization. In: International Conference on Computational Science ({ICCS})
  2018, pp. 90--102. Springer (2018). \doi{10.1007/978-3-319-93701-4_7}

\bibitem{robinson2002particle}
Robinson, J., Sinton, D., Rahmat-Samii, Y.: Particle swarm, genetic algorithm,
  and their hybrids: Optimization of a profiled corrugated horn antenna. In:
  2002 {IEEE} Antennas and Propagation Society International Symposium ({IEEE}
  Cat. No.02CH37313). vol.~1, pp. 314--317. IEEE (2002)

\bibitem{sengupta2019particle}
Sengupta, S., Basak, S., Peters, R.A.: {Particle} {Swarm} {Optimization}: {A}
  survey of historical and recent developments with hybridization perspectives.
  Machine Learning and Knowledge Extraction  \textbf{1}(1),  157--191 (2019)

\bibitem{Shao2023PGA}
Shao, K., Song, Y., Wang, B.: {PGA}: {A} new hybrid {PSO} and {GA} method for
  task scheduling with deadline constraints in distributed computing.
  Mathematics  \textbf{11}, ~1548 (2023). \doi{10.3390/math11061548}

\bibitem{shi2003hybrid}
Shi, X., Lu, Y., Zhou, C., Lee, H., Liang, Y.: Hybrid evolutionary algorithms
  based on {PSO} and {GA}. In: Congress on Evolutionary Computation ({CEC}).
  pp. 2393--2399. IEEE (2003). \doi{10.1109/CEC.2003.1299387}

\bibitem{shi2003pso}
Shi, X., Wan, L., Lee, H., Yang, X., Wang, L., Liang, Y.: {PSO-GA} based hybrid
  evolutionary algorithm. In: Proceedings of the International Conference on
  Machine Learning and Cybernetics. pp. 1735--1740. IEEE (2003)

\bibitem{Simaiya2024}
Simaiya, S., Lilhore, U.K., Sharma, Y.K., Rao, K.B.V.B., Maheswara~Rao, V.V.R.,
  Baliyan, A., Bijalwan, A., Alroobaea, R.: A hybrid cloud load balancing and
  host utilization prediction method using deep learning and optimization
  techniques. Scientific Reports  \textbf{14}(1), ~1337 (2024).
  \doi{10.1038/s41598-024-51466-0}

\bibitem{thangaraj2011particle}
Thangaraj, R., Pant, M., Abraham, A., Bouvry, P.: Particle swarm optimization:
  {Hybridization} perspectives and experimental illustrations. Applied
  Mathematics and Computation  \textbf{217}(13),  5208--5226 (2011)

\bibitem{DBLP:journals/jocs/TurekSKAPBK16}
Turek, W., Stypka, J., Krzywicki, D., Anielski, P., Pietak, K., Byrski, A.,
  Kisiel{-}Dorohinicki, M.: Highly scalable erlang framework for agent-based
  metaheuristic computing. J. Comput. Sci.  \textbf{17},  234--248 (2016).
  \doi{10.1016/J.JOCS.2016.03.003}

\bibitem{yang2007hybrid}
Yang, B., Chen, Y., Zhao, Z.: A hybrid evolutionary algorithm by combination of
  {PSO} and {GA} for unconstrained and constrained optimization problems. In:
  Proceedings of the IEEE International Conference on Control and Automation.
  pp. 166--170 (2007). \doi{10.1109/ICCA.2007.4376340}

\end{thebibliography}

\end{document}